\pdfoutput=1 
\documentclass{article}

\usepackage{microtype}
\usepackage{graphicx}
\usepackage{subfigure}
\usepackage{booktabs} 

\usepackage{hyperref}

\usepackage[accepted]{icml2025}

\usepackage{amsmath}
\usepackage{amssymb}
\usepackage{mathtools}
\usepackage{amsthm}
\usepackage{dsfont}
\usepackage[most]{tcolorbox}

\usepackage[capitalize,noabbrev]{cleveref}

\definecolor{blue}{HTML}{4073b5} 

\theoremstyle{plain}

\theoremstyle{definition}

\theoremstyle{remark}

\newcommand{\leo}[2][]{\todo[inline, color=blue!20, #1]{Leo: #2}}
\newcommand{\tom}[2][]{\todo[inline, color=red!20, #1]{Tom: #2}}

\usepackage{enumitem}
\usepackage{xcolor}
\usepackage{colortbl}
\definecolor{lightgray}{gray}{0.85}
\definecolor{darkgray}{gray}{0.7}
\definecolor{customblue}{rgb}{0.27, 0.36, 0.55}

\usepackage[disable,textsize=tiny]{todonotes}
\usepackage{array}
\usepackage{makecell}

\usepackage{amsmath,amsfonts,bm}

\def\eqref#1{equation~\ref{#1}}

\def\1{\bm{1}}

\DeclareMathAlphabet{\mathsfit}{\encodingdefault}{\sfdefault}{m}{sl}
\SetMathAlphabet{\mathsfit}{bold}{\encodingdefault}{\sfdefault}{bx}{n}

\def\gC{{\mathcal{C}}}

\def\gP{{\mathcal{P}}}

\def\gX{{\mathcal{X}}}
\def\gY{{\mathcal{Y}}}

\def\sR{{\mathbb{R}}}

\newlength{\myboxwidth}
\icmltitlerunning{Toward Meaningful Progress in Adversarial Alignment for LLMs}

\begin{document}

\twocolumn[

\icmltitle{\texorpdfstring{Adversarial Alignment for LLMs Requires\\ Simpler, Reproducible, and More Measurable Objectives}{Adversarial Alignment for LLMs Requires Simpler, Reproducible, and More Measurable Objectives}}



\icmlsetsymbol{equal}{*}

\begin{icmlauthorlist}
\icmlauthor{Leo Schwinn}{tum}
\icmlauthor{Yan Scholten}{tum}
\icmlauthor{Tom Wollschl\"ager}{tum}
\icmlauthor{Sophie Xhonneux}{mila,montreal}
\icmlauthor{Stephen Casper}{mit}
\icmlauthor{Stephan G\"unnemann}{tum}
\icmlauthor{Gauthier Gidel}{mila,montreal,cifar}
\end{icmlauthorlist}

\icmlaffiliation{tum}{Technical University of Munich}
\icmlaffiliation{mila}{Mila}
\icmlaffiliation{montreal}{Universit\'{e} de Montr\'{e}al}
\icmlaffiliation{cifar}{Canada AI CIFAR Chair}
\icmlaffiliation{mit}{MIT}

\icmlcorrespondingauthor{}{adversarial-alignment-position@googlegroups.com}

\icmlkeywords{Machine Learning, ICML}

\vskip 0.3in
]



\printAffiliationsAndNotice{} 





\begin{abstract}
Misaligned research objectives have considerably hindered progress in adversarial robustness research over the past decade.
For instance, an extensive focus on optimizing target metrics, while neglecting rigorous standardized evaluation, has led researchers to pursue ad-hoc heuristic defenses that were seemingly effective. Yet, most of these were exposed as flawed by subsequent evaluations, ultimately contributing little measurable progress to the field.
In this position paper, we illustrate that current research on the robustness of large language models (LLMs) risks repeating past patterns with potentially worsened real-world implications. To address this, we argue that \textbf{realigned objectives are necessary for meaningful progress in adversarial alignment.}
To this end, we build on established cybersecurity taxonomy to formally define differences between past and emerging threat models that apply to LLMs.
Using this framework, we illustrate that progress requires disentangling adversarial alignment into addressable sub-problems and returning to core academic principles, such as measureability, reproducibility, and comparability.
Although the field presents significant challenges, the fresh start on adversarial robustness offers the unique opportunity to build on past experience while avoiding previous mistakes.

\end{abstract}



\section{Introduction}

\leo{Title suggestions: 

Adversarial Alignment of LLMs Requires Less Complex and More Measureable Objectives

Adversarial Alignment of LLMs Requires Simpler and More Measurable Objectives

Adversarial Alignment of LLMs Requires Disentangled and More Measurable Objectives
}

\tom{more title suggestions:

unspecific:

A way forward for adversarial alignment 

How to approach adversarial alignment

}

\leo{Add proposal to the beginning of the paper that we will reevaluate the state of research in a year and that  other researchers are welcome to be included if they offer arguments for / against the position and how things changed}


Security risks in computer science have been a prevalent issue for decades~\cite{valiant_learning_1985, kearns_learning_1993}. 
This ongoing challenge has resulted in an ``arms race" that includes the continued development of new attacks, such as malware and phishing, as well as defense mechanisms. 

In the field of deep learning, \citet{szegedy_intriguing_2014} discovered that deep neural networks are highly susceptible to \textit{adversarial examples} -- input perturbations optimized to mislead models into making predictions that are erroneous or misaligned with their intended behavior. In response, countless defense strategies were proposed to safeguard neural networks against these attacks in the last decade. Yet, a majority of newly proposed heuristic defenses were eventually exposed as flawed by subsequent evaluations, often by using standard attack protocols that already existed at the time of the defense publication~\citep{tramer_adaptive_2020}. 

We investigate the research state of adversarial alignment in large language models (LLMs), which we define as the ability of an aligned model to maintain its intended training objective despite input perturbations. %
We observe that adversarial alignment risks following the same cycle of flawed defenses and subsequent rectified evaluations seen in past adversarial robustness research but with considerably amplified challenges and stakes~\cite{hendrycks2022x, zou2023universal, schwinn2023adversarial}. 
Unlike previous robustness research that generally focused on well-defined problems like image classification~\cite{goodfellow_explaining_2015}, assessing LLM capabilities is considerably more challenging due to inherent ambiguities of the alignment problem and the complexity of natural language~\cite{wolf2023fundamental, andriushchenko2024jailbreaking, li2024llm}. Furthermore, the potential harm associated with LLMs is substantially greater due to their advanced capabilities and widespread availability~\cite{hendrycks2022x} in particular as they start being used as autonomous agents.  



\begin{table*}[ht]
    \centering
    \caption{Non-exhaustive comparison of robustness research in previous domains and LLMs.}
     \newcommand{\cellwidth}{2.15cm}
     \newcommand{\linesp}{3pt}
     \newcommand{\cbox}[1]{\parbox[t][0.77cm][t]{\cellwidth}{\centering #1}}
    \tiny	
\begin{tabular}{l*{3}{>{\centering\arraybackslash}m{\cellwidth}}*{3}{>{\centering\arraybackslash}m{\cellwidth}}}
    \addlinespace
     & \multicolumn{3}{@{\hspace{3mm}}c@{\hspace{3mm}}}{\cellcolor{customblue}\textcolor{white}{\textbf{Attack goals}}} & \multicolumn{3}{@{\hspace{3mm}}c@{\hspace{3mm}}}{\cellcolor{customblue}\textcolor{white}{\textbf{Attack Capabilities}}} \\
    \addlinespace[\linesp]
    \rowcolor{darkgray} & \textbf{Objectives} & \textbf{Attacks} & \textbf{Datasets} & \textbf{Access} & \textbf{Constraints} & \textbf{Frameworks} \\
    \rowcolor{lightgray}\textbf{Previous} & \cbox{Clear objectives\\(e.g., classification)\\\cite{szegedy_intriguing_2014}} & \cbox{Generally reliable\\\cite{croce2020reliable}} & \cbox{Standardized\\\cite{croce2020robustbench}} & \cbox{Generally white-box\\and open-source\\\cite{croce2020robustbench}} & \cbox{Tractable but incomplete\\(e.g., $\ell_p$)\\\cite{szegedy_intriguing_2014}} & \cbox{Standardized\\\cite{croce2020robustbench}} \\
    \addlinespace[\linesp]
    \rowcolor{lightgray}\textbf{LLM} & \cbox{Entangled objective of\\alignment \& robustness\\\cite{zou2023universal}} & \cbox{Currently weak\\\cite{li2024llm}} & \cbox{Entangled notions of\\harmfulness\\\cite{mazeika2024harmbench}} & \cbox{Often black-box and\\proprietary models\\\cite{zou2023universal}} & \cbox{No constraints\\\cite{chao2023jailbreaking,mazeika2024harmbench}} & \cbox{Varying evaluation settings\\\cite{mazeika2024harmbench}} \\
\end{tabular}
    \label{tab:comparison}
\end{table*}

We argue that the past lack of progress can be largely attributed to a narrow focus on improving benchmark numbers without sufficient attention to rigorous evaluations and clear evaluation criteria. This led to the proliferation of ad-hoc defenses that relied on security through obscurity and ultimately proved ineffective, thus not providing a solid foundation for future work. In this context, the field of adversarial alignment risks ``treading water,'' expending significant effort but failing to make meaningful progress.  

\newpage
Our position on adversarial alignment in LLMs is as follows:

\begin{tcolorbox}[
    colback=white, colframe=customblue, coltitle=white, fonttitle=\bfseries, 
    rounded corners, enhanced, 
    title=Position, 
    attach boxed title to top left={yshift=-2mm, xshift=5mm}, 
    boxed title style={colback=customblue, rounded corners},
    boxsep=0.5mm
]
 We argue that researchers deal with:\,\,\textbf{1)} vague problem definitions, where alignment and robustness are inherently entangled in the adversarial alignment problem, \textbf{2)} complex and non-reproducible evaluations, and \textbf{3)}  an emphasis on state-of-the-art attack performance against proprietary models over reproducible and comparable open-source research.\\ 
 \textbf{Thus, meaningful progress in adversarial alignment for LLMs requires simpler, reproducible, and more measurable objectives.}
\end{tcolorbox}

Our main contributions are:


\begin{itemize}[noitemsep,nolistsep,topsep=-2pt,leftmargin=0.6cm] 
    \item We systematically identify challenges in previous robustness research, how they apply to LLMs, and how new challenges emerged. Based on this analysis: 
    \item We demonstrate how adversarial alignment intertwines the challenges of alignment and robustness, making robustness evaluation difficult, as it inherits the challenges of measuring alignment, such as ambiguity in success criteria. 
    While acknowledging the importance of comprehensive assessment, we advocate for a complementary approach: focusing on measurable sub-problems with more measureable objectives, where not every work aims to address the complete picture.
    
    \item Towards the same goal of improving measurability, we propose simplifying robustness benchmarks by evaluating specific types of harm individually rather than combining complex concerns like copyright infringement, fairness, and toxicity into a single evaluation.
    
    \item We emphasize the need for academia to prioritize reproducible research over chasing SOTA performance on proprietary models. In this context, we propose fostering open-source research with accessible models. 
    \item Computational overhead, vast number of hyperparameters, and varying implementation details hinder comparability between different works. We advocate for a practical approach to improve reproducibility and comparability: community-driven leaderboards and standardized benchmarks to encourage best practices in adversarial robustness research.  

\end{itemize}

\section{Structure of Our Argument}\label{sec:structure}

We structure our position as follows: We define a taxonomy of adversarial robustness threats based on common cybersecurity frameworks including \textbf{I)} robustness goals and \textbf{II)} adversary capabilities. For both elements of this taxonomy, we follow a parallel argument structure: We \textbf{A)} define past and present threat models (\S\ref{sec:pos-goals-definition} and \S\ref{sec:pos-capability-definition}), \textbf{B)} discuss historical challenges and connections to upcoming and exacerbated issues in the LLM domain (\S\ref{sec:pos-goals-challenges}, and \S\ref{sec:pos-capability-challenges}), and \textbf{C)} explore the applicability of past insights to new problems and how current research objectives can be realigned to promote measurable progress (\S\ref{sec:pos-goals-realigned} and \S\ref{sec:pos-capability-realigned}). We provide a non-exhaustive comparison between past and current robustness research in Table~\ref{tab:comparison}.


\section{Related Positions and Future Outlook} 

\textbf{Concurrent positions.} Our findings are reinforced by the concurrent work of~\citet{rando2025adversarial}, who independently identify similar fundamental challenges in adversarial alignment of LLMs. 
Their research offers detailed case studies of specific threat models like jailbreaks, poisoning, and unlearning. While we investigate specific failures as well--albeit in less detail--, our focus is on developing a high-level taxonomy and proposing forward-looking solutions to emerging challenges. 
Together, both works offer complementary perspectives toward a more comprehensive understanding of the identified issues.

\textbf{Call for positions.} We will revisit and evaluate the arguments presented in this position paper one year or more from its publication. This retrospective analysis will examine whether our identified challenges and proposed solutions have proven relevant regarding the field's progress in adversarial robustness for LLMs. We warmly welcome other researchers to contribute their perspectives and collaborate on the retrospective, whether they want to support or challenge the positions outlined in the remainder of this paper. Through this collaborative reflection, we aim to maintain accountability for our arguments.



\vspace{-10pt}
\section{Adversarial Robustness Taxonomy}

To systematically define differences between prior adversarial robustness research and emerging threat models in LLMs we introduce a taxonomy inspired by established cybersecurity definitions organized into robustness goals and capabilities~\citep{papernot_practical_2017, carlini_evaluating_2019}. In the remainder of this work, we use this taxonomy to formally define differences between previous threat models and emerging ones in LLMs.

\vspace{-5pt}
\subsection{Adversarial Robustness Goals}\label{sec:tax-goals}%
The overarching goal of adversarial robustness research is to improve the robustness of neural networks against input perturbations. To this end, clearly defining this goal is crucial for consistent and meaningful comparisons of robustness evaluations across multiple works. 
This requires introducing a sound mathematical framework that specifies the conditions under which robustness should be achieved. 
For example, considering evasion (test-time) robustness, the goal is to determine the worst-case change in the model's output under constrained changes in the input data:
\begin{equation}\label{eq:adv-goal-evasion}
    \delta^*(x) = \max_{\tilde{x} \in \gC(x)} \; d_{out} (f(x), f(\tilde{x})), 
\end{equation}
where $x$ is a fixed input, $f$ is the model, $d_{out}$ is a general distance metric between the model outputs, and $\gC(x)$ is the set of possible perturbations within a certain distance of $x$ (see \S\ref{sec:tax-capability}). 
%
%
The optimization problem in \autoref{eq:adv-goal-evasion} establishes a fundamental definition of adversarial robustness in evasion settings, and serves as the foundation for its evaluation. Solving this optimization problem is NP-hard for most practical scenarios~\cite{katz2017reluplex}. To address this challenge, we require algorithms to assess robustness by computing lower and upper bounds on the model's robustness $\delta^*$. 
In the evasion example, lower bounds $\underline{\delta}(x)$ on $\delta^*(x)$ defined in \autoref{eq:adv-goal-evasion} can be derived using adversarial attacks, and upper bounds $\overline{\delta}(x)$ can be derived using robustness certification. Together, this allows to estimate the robustness of a given model: $\underline{\delta}(x) \leq \delta^*(x) \leq \overline{\delta}(x)$. In general, attacks and certificates provide (provable) bounds on robustness, and defenses (such as adversarial training) constitute strategies for achieving the goal of improved robustness.
\vspace{-5pt}
\subsection{Capability in Adversarial Robustness}\label{sec:tax-capability}%
We define capabilities of robustness algorithms by 1) their knowledge of the optimization problem, e.g., about the underlying functions $f$, 2) possible constraints, e.g., perturbation budgets, and 3) the required computational effort. 

\textbf{Knowledge.} Knowledge and access classifications typically include up to four categories, consisting of white-box, gray-box, black-box, and no-box ranging from full understanding and complete access to the model and its defenses (white-box) to no knowledge and no direct access (no-box), with varying degrees of knowledge and access in between (gray-box, black-box)~\citep{papernot_practical_2017, bose_adversarial_2020}.

The white-box threat model has emerged as the most common evaluation scenario in the adversarial robustness literature, as it enables the strongest attacks, thereby providing the most accurate quantification of the robustness of a respective defense~\citep{papernot_practical_2017}.
This aligns with Kerckhoff's Principle, which asserts that a system's security should not depend on obscurity, i.e., the secrecy of its design or implementation. Relying on obscurity introduces vulnerabilities, as once that obscurity is compromised, the entire system's robustness is at risk~\citep{sasa_kerk_2008,athalye_obfuscated_2018}.
The widespread adoption of the white-box threat model has been enabled through the open sourcing of newly published defenses and models.

\textbf{Perturbation constraints.} Constraints on adversarial perturbations serve two essential roles in robustness assessment: First, they reflect practical limitations, such as maintaining valid input domains (e.g., pixel values within image bounds) or ensuring malicious perturbations remain undetected. Second, they provide meaningful evaluation settings, as unconstrained adversaries can typically bypass any defense mechanism, making such scenarios more suitable as sanity checks than realistic threat models~\cite{goodfellow_explaining_2015}.
Formally, perturbation constraints $\gC$ are typically defined by bounding a distance metric between the original and perturbed inputs, $d_{in}(x, x') \leq \epsilon$. This formalization enables a consistent framework for evaluating and comparing robustness under defined conditions \cite{carlini_evaluating_2019}.

\textbf{Computational effort \& complexity.} We extend the attack capability definition to include practicality constraints on attacks and defenses, such as computational effort or pipeline complexity. Practicality constraints enable more realistic modeling of real-world attacks by reflecting the costs associated with both attacks and defenses. Cybersecurity threat models often assume an inverse relationship between these costs: a higher cost for an attacker, typically means a lower cost for the defender~\cite{barreno_security_2010}.

\vspace{-10pt}
\section{Position on Robustness Goals}


\subsection{Robustness Goals in Previous Settings}\label{sec:pos-goals-definition}

Over the past decade, the primary goal in robustness research has been to define and improve robustness of single-output models $f\,$$:$$\, \gX \rightarrow \gY$ for (e.g.\ image) classification ($\gY = \{1, \ldots, C\}$) or regression tasks ($\gY=\sR$), as well as of multi-output models $f : \gX \rightarrow \gY^d$, extending these tasks to higher-dimensional output spaces (e.g., graph-structured data).
Robustness in the output space $d_{out}$ has been directly assessed using indicator functions $\mathds{1}[y \neq y']$ for classification or $\ell_p$ norms for regression.

\vspace{-7pt}
\subsubsection{Robustness Goals in LLMs}\label{sec:pos-goals-definition-llm}
Unlike traditional robustness settings that focus on simple single- or multi-output functions, large language models are generative models and describe complex output distributions in the natural language domain. They belong to a much broader class of parameterized functions $f_\theta: V^* \rightarrow \gP(R)$ that, given a vocabulary $V$, map texts of arbitrary length $V^*$ to distributions $\gP(R)$ over possible responses~$R$ and can only be evaluated sequentially: 
$$f_\theta(y_1, \ldots, y_T | x) = \prod_{t=1}^T f_\theta(y_t | y_{t-1}, \ldots, y_1, x),$$ 
where $f_\theta(y_t| y_{t-1}, \ldots, y_1, x)$ is the conditional probability over the next possible tokens $y_t\in V$ given the previous tokens $y_{t-1}, \ldots, y_1$ and the input $x$. 


Notably, this fundamental difference complicates defining robustness goals in the output space. For example, an attack targeting a model's safety alignment must account for: 1) an exponential arbitrary-length outputspace, 2) subjective definitions of success, as human raters may disagree on what constitutes harmfulness, 3) ambiguous edge cases, such as the gray area between outright refusal, incorrect but affirmative responses, and explicitly harmful content, and 4) the inability to rely on well-defined distance metrics, such as $\ell_p$ norms or edit distance, which fail to capture semantic meaning, necessitating subjective human evaluation or auxiliary models~\cite{zou2023universal, mazeika2024harmbench}. 

\vspace{-5pt}
\subsection{Previous and New Challenges}\label{sec:pos-goals-challenges}

\textbf{Best practices.} Despite well-defined threat models and a focus on relatively simple architectures and datasets, many early robustness claims proved over-optimistic. This could be largely attributed to faulty evaluation practices, where defenses often relied on security through obscurity~\cite{athalye_obfuscated_2018}. To address this, researchers established comprehensive evaluation guidelines. Several studies demonstrated that adhering to a few key principles -- such as computing gradients using expectation over random transformations or approximating non-differentiable components through differentiable approximation -- was enough to break most defenses and substantially improve the reliability of evaluations~\cite{athalye_obfuscated_2018, uesato2018adversarial}. 

Some early works already show that simple modifications to attack strategies can still break seemingly robust state-of-the-art LLM defenses~\cite{schwinn2024revisiting, thompson2024flrt}. While fundamental principles against security through obscurity remain relevant~\cite{athalye_obfuscated_2018}, evaluations are generally more difficult. Simply following previous best practices appears to be generally insufficient in LLMs, and adaptive evaluation strategies entail significant human effort~\cite{andriushchenko2024jailbreaking}. 

\looseness=-1
\textbf{Sanity checks.} To identify errors in robustness assessments, simple sanity checks could be performed to identify faulty evaluations~\cite{uesato2018adversarial, carlini_evaluating_2019}. This included verifying that attacks with larger computational budgets consistently achieve better results and that defenses fail when attacks are provided with an unconstrained perturbation budget.

However, sanity checks in previous work with continuous input domains are difficult to apply to the LLM assistant setting and may require rethinking~\citep{athalye_obfuscated_2018, carlini_evaluating_2019, tramer_adaptive_2020}. Previously, an unconstrained attack budget would generally lead to a successful attack in the vision domain, e.g., an attacker could replace an image with another image to fool a classifier~\cite{carlini_evaluating_2019}. Testing whether a defense fails in this threat model was one way to verify the validity of the evaluation. Similarly, defenses that remained robust under this threat model could be identified as relying on security through obscurity~\citep{athalye_obfuscated_2018}. In the LLM setting, these sanity checks cannot be applied directly as even unconstrained attacks are currently not able to break models without considerable manual effort~\cite{li2024llm}.

\textbf{Certified approaches.} Robustness certification provides a mathematically rigorous path to measuring and improving model robustness. The combination of clear threat models and relatively small model sizes enabled various verification approaches, from closed-form certificates~\cite{tjeng2017evaluating}, to principled approximations~\cite{gowal2018effectiveness}, and probabilistic bounds~\cite{cohen2019certified}. 

Yet, most certification approaches could not be scaled to large datasets~\cite{tjeng2017evaluating}, or introduced substantial computational overheads~\cite{cohen2019certified}. Moreover, a significant gap exists between the robustness bounds provided by certificates and attacks.
This issue is further amplified for LLMs, due to their model size, combined with the complexity of the natural language space and creates major challenges for certified robustness. Thus, ensuring the reliability of empirical evaluations becomes even more~crucial. 
\vspace{-15pt}
\subsubsection{Emerging Problems}\label{sec:goal-amplified}



\textbf{Vague robustness definitions.} The generative nature and natural language domain of LLMs make it considerably more difficult to define what constitutes a successful attack and evaluate the robustness goal, as it involves comparing not just single outputs but entire distributions of possible responses~\cite{scholten2025a}. Furthermore, a lack of consensus on desired model behavior and a vague definition of robustness introduce further challenges in evaluating robustness. Different language models are aligned to distinct objectives, making it challenging to ensure current benchmarks measure comparable underlying properties across models. For example, some existing robustness benchmarks evaluate copyright compliance as a security property~\cite{mazeika2024harmbench}. However, this assessment may be misaligned with model design: some models are not explicitly constrained against reproducing existing content and may generate verbatim copies of text or code even in standard, non-adversarial contexts. Since overall robustness is typically reported as an aggregate score, these evaluations risk conflating model robustness properties (i.e., the stability of a model output under attack) with behaviors that simply reflect different design choices in model alignment. 

Moreover, defining what constitutes unwanted, e.g., harmful behavior, is context-dependent and varies considerably based on cultural background, personal values, and other factors. This ambiguity makes it difficult to create universally agreed-upon robustness goals and benchmarks to measure robustness against such harms. Early evaluations in LLM robustness relied on simple string matching to identify benign and harmful outputs~\cite{zou2023universal} or human labeling~\cite{zhu2023autodan}. Later works mostly rely on LLMs as judges to classify generated sequences~\cite{zhu2023autodan, mazeika2024harmbench}, making their results dependent on the capability of the LLM judge, which are often only slightly more accurate than random guessing~\cite{tan2024judgebench, shi2024judging}. Further, results may be difficult to compare as proprietary models are often used as judges, which may change over time~\cite{zhu2023autodan}. Consequently, evaluating if the robustness goal has been achieved involves considerable overhead in the natural language domain. 
In contrast, previous robustness evaluations in classification tasks can be evaluated more objectively using standard classification datasets, where stability is measured by analyzing how model predictions respond to perturbations of the input.

\textbf{Weak automatic adversarial attacks.} In prior robustness research, white-box attacks often served as a reliable robustness estimate and were generally able to completely break undefended models regardless of model scale~\cite{shao2021adversarial}. However, this assumption does not hold for LLMs. As attacks must be optimized in a discrete natural language space, standard gradient-based methods are less effective. Here, the optimization problem becomes a complex combinatorial search with an attack space that grows exponentially with sequence length. As a consequence, current automated adversarial attack algorithms against LLMs are still outperformed by human jailbreakers~\cite{li2024llm} and typically achieve considerably less than 100\% attack success rate (ASR) even against undefended models~\cite{zou2023universal}. The weaknesses of automatic attacks compared to human jailbreaks and the stark contrast to robustness results in previous domains suggest that current attacks overestimate robustness. This is supported by~\citet{andriushchenko2024jailbreaking}, who achieved near 100\% success rates by combining manually human-designed attacks with optimization algorithms. While there are early efforts towards more efficient automatic white-box evaluations in the latent space~\cite{schwinn2023adversarial, schwinn2024soft, che2025model}, these threat models may be viewed as unrealistic~\cite{zou2024improving}, as they require white-box model access and are therefore mostly relevant as a development tool~\cite{che2025model}.

\leo{Include discussion about diverse metrics, efficiency, perplexity, transferability, robustness-accuracy trade-off}

\leo{Add sth about ASR not capturing everything in LLMs}

\vspace{-5pt}
\subsection{Realigned Objectives}\label{sec:pos-goals-realigned}

We identified two major issues related to robustness goal definitions in the context of LLMs. First, vague and complex robustness definitions depending on semantic outputs make it difficult to measure robustness. Secondly, currently, weak attacks give no reliable estimates of model robustness.  

\textbf{A Case for Simpler Objectives.} We argue that both these issues can be addressed by disentangling problem definitions and reducing overall complexity in evaluation pipelines. Recent works generally measure the performance of a defense or attack by assessing the harmfulness of the respective LLM generations. This approach necessitates complex pipelines involving: \textbf{A)} the collection of diverse data to measure alignment or harmfulness, \textbf{B)} optimization of the attack using proxy objectives, such as maximizing the likelihood of harmful completions to manipulate the model into complying with harmful requests, and \textbf{C)} using an LLM judge to classify the harm of the generated output. While this approach mimics real-world threat models, we believe the multitude of design decisions within these complex pipelines adds unknown biases to evaluations and leads to low reproducibility, comparability, and measurability. Even without the additional complications that come with LLMs, measuring progress reliably in machine learning has been a major challenge due to insufficient reporting of results~\cite{dodge2019show}, transferability of results~\cite{liao2021we}, lack of statistical significance in improvements~\cite{card2020little}, and dataset leakage~\cite{kapoor2022leakage}, which provides a clear argument for reducing complexity. 

The entanglement of different notions of robustness (e.g., harmfulness vs. copyright violations) complicates accurate assessment of model robustness. 
Furthermore, small changes in some parts of the pipeline, such as a different version of a proprietary LLM judge, can drastically alter robustness evaluations and thus harm reproducibility and comparability. 
We propose a realignment of research objectives, moving away from tackling the complex adversarial alignment problem as a whole in \textit{every single work}. Instead, we propose a \textit{complementary} approach to address individual, well-defined sub-problems in isolation. This approach is motivated by the assumption that current weaknesses in adversarial attacks on LLMs primarily stem from limitations of optimization algorithms~\cite{carlini2023aligned}. Improving the capability of these algorithms, however, is largely independent of the complexities of the alignment problem and can be achieved more effectively using simpler and reproducible benchmarks. Thus, we propose the following:

\textbf{Simpler datasets.} Addressing \textbf{A)}, rather than creating large, diverse datasets that cover diverse aspects of alignment, we suggest focusing on narrowly defined and simple threat settings (e.g., explicit instructions to use slurs). While such datasets may not fully reflect model alignment in real-world deployment, they offer two major benefits. 
Firstly, they enable reliable assessments of model \textit{robustness}. Detecting slurs in generated text, for instance, is significantly easier than judging harmfulness, closing the gap to well-defined robustness goals in previous robustness research (see \S\ref{sec:pos-goals-definition}). By disentangling robustness from alignment, this approach reduces ambiguity, facilitating measurable research progress. Furthermore, disentanglement reduces evaluation pipeline complexity (e.g., by removing the necessity to judge model outputs), which lowers computational overhead and minimizes the risk of evaluation errors. 
Secondly, these datasets reduce the necessity for researchers to engage with a large spectrum of harmful or toxic material. Moreover, they eliminate uncertainties that stem from cultural and personal background differences in evaluating the robustness objective. Here, benchmarks targeting different alignment properties can be released iteratively, allowing for more nuanced evaluations without compromising measureability.

\textbf{Measureable proxy objectives.} Addressing \textbf{B)} and \textbf{C)}, rather than assessing model generations directly, attack algorithms can be evaluated based on their ability to optimize predefined objectives. 
For example, an attack might aim to increase the probability of an affirmative response to a harmful request or achieve a specific benign target. This disentangles attack optimization from alignment evaluation and enables more direct comparisons of algorithmic efficiency, as key metrics can be tracked throughout the attack. Directly evaluating proxy metrics additionally simplifies evaluation pipelines by removing the need for an LLM judge. 

A concrete example of such a proxy objective could be a model fine-tuned to withhold responses to specific requests (e.g., to reveal a key) unless provided with a hidden, predetermined input (similar to model backdoors). In this setup, the goal is to enforce a behavioral restriction that makes the model non-responsive unless the "trigger" is found. This approach not only defines a clear optimization target that is known a priori to be theoretically achievable by the attack but also isolates the adversarial optimization from the alignment objective. While it may be difficult for academics to design proxy objectives that mimic real-world defenses (e.g., due to a lack of knowledge regarding proprietary safety mechanisms), this may be resolved by companies providing dedicated proxy objectives, such as models hiding a key.

\begin{tcolorbox}[
    colback=white, colframe=purple!80!black, coltitle=white, fonttitle=\bfseries, 
    rounded corners, enhanced, 
    title=Key Takeaways, 
    attach boxed title to top left={yshift=-2mm, xshift=5mm}, 
    boxed title style={colback=purple!80!black, rounded corners},
    boxsep=0.5mm,
    left=2mm, 
    right=2mm
]
Adversarial alignment entangles the alignment and robustness problem. By reducing complexity in goal specifications and benchmarks, we can isolate manageable sub-problems with verifiable objectives. This improves reproducibility, comparability, and measurability, and thereby facilitates reliable progress.
\end{tcolorbox}

\vspace{-10pt}
\section{Position on Attacks Capabilities}

\subsection{Attack Capability in Previous Settings}\label{sec:pos-capability-definition}
\textbf{Attacker knowledge.} As detailed in \S\ref{sec:tax-capability}, attacker knowledge of the model $f$ and the underlying defense mechanisms can range from black-box to white-box scenarios, each with varying levels of access. Past works generally recommended evaluating defenses in the white-box threat model due to multiple reasons~\cite{athalye_obfuscated_2018}. 

First, white-box attacks provide the most accurate upper bound on defense robustness. We argue that adversarial robustness research primarily aims to uncover fundamental insights for long-term improvements in neural network robustness rather than focusing on immediate real-world system protection; this threat model is most informative. Secondly, in line with Kerckhoff’s Principle (see \S~\ref{sec:tax-capability}), defense through obscurity alone is generally discouraged. This principle has proven especially relevant in machine learning, where defenses based on obscurity have repeatedly been shown to be unreliable~\cite{athalye_obfuscated_2018}.

\textbf{Perturbation constraints.} Typically, input distances $d_{in}$ have been measured using $\ell_p$ norms in continuous settings and discrete metrics such as edit distance in discrete settings. These well-defined metrics allowed for tractable enforcement of constraints and facilitated reproducible, comparable threat models across studies.
\vspace{-5pt}
\subsubsection{The Novel LLM Setting}\label{sec:capability-llm-setting}






\textbf{Attacker knowledge.}  Unlike past research where white-box evaluations were the norm, proprietary LLMs are only accessible in black-box settings with limited control over their exact hyperparameter settings (e.g., in most cases, it is not possible to force deterministic generations). While many publicly available models are accessible, recent works often perform their evaluations on proprietary models due to their real-world relevance (e.g., widespread use) and state-of-the-art capabilities~\cite{andriushchenko2024jailbreaking}. Moreover, researchers not only lack access to proprietary models but even for open-weight models face limited knowledge about the training datasets and specific safety measures.

\textbf{Perturbation constraints.} Unlike distance-bound metrics in previous robustness research, LLM threat models allow unrestricted perturbations in input distance $d_{in}$, with the only constraint being that modifications must occur in natural language through discrete changes, as attacks typically target models via APIs without internal layer access.

First works have begun exploring semantic constraints on adversarial text, focusing on attacks that maintain high text quality (measured by metrics like perplexity) or avoid detection as harmful content~\cite{liu2023autodan}.  
\vspace{-5pt}
\subsection{Previous and New Challenges}\label{sec:pos-capability-challenges}


\textbf{Third-party evaluations.} Many defenses against adversarial attacks proposed in earlier research were later found to be ineffective. Numerous studies have demonstrated this by developing adaptive attacks that successfully bypassed over $50$ defenses published in top machine learning venues~\cite{uesato2018adversarial, athalye_obfuscated_2018, tramer_adaptive_2020}. We argue that, while we believe adaptive attacks are still crucial for reliable evaluations, a similar pattern is unlikely to emerge in LLM adversarial robustness research. The high computational cost of running these models, combined with the inefficiency of current attacks, makes breaking them significantly more time- and resource-intensive~\cite{andriushchenko2024jailbreaking}. Thus, many researchers will lack the resources or incentive to conduct third-party evaluations.

\textbf{Public leaderboards and common frameworks.} To address faulty evaluations, the community established public leaderboards, where models were evaluated within the same framework. These leaderboards emphasized reliable defense strategies while prohibiting approaches that were difficult to evaluate due to obscurity (e.g., approaches that included non-differential components or randomness). Frameworks included predefined attack settings and implementations and thus enabled reliable evaluations and comparable, reproducible research. However, LLM evaluations are difficult to compare due to the numerous implementation-specific variations outlined above and current works mostly transfer and adapt attack implementations from original works to their own codebases~\cite{xhonneux2024efficient}. 

\vspace{-5pt}
\subsubsection{Emerging Problems}
\textbf{Evaluations on proprietary models.} Black-box access reduces the reliability of model evaluations, making it more difficult to compute the worst-case model robustness as formalized in~\autoref{eq:adv-goal-evasion}. Since the exact specifications of proprietary APIs are unknown, it also remains unclear if robustness increases are due to model improvements, additional filtering steps, or other measures that providers deploy. For example,~\citet{zou2023universal} initially demonstrated that even the largest proprietary models are vulnerable to transfer attacks optimized on open-source models. However, some proprietary models were updated to be robust against these attacks shortly afterward~\cite{hurst2024gpt}. It now remains unclear if this robustness is due to \textit{obscurity} or more fundamental improvements in robustifying LLM. 
Moreover, constantly changing model versions hinder reproducibility and comparability, as even when versions are reported, their availability for evaluation remains uncertain.

\textbf{Computational effort \& complexity.} Adversarial robustness research has traditionally focused on small datasets and models due to the problem's difficulty~\cite{bartoldson2024adversarial}. In contrast, studying robustness in LLMs demands far greater resources, with even the smallest benchmarked models exceeding 7 billion parameters~\cite{xhonneux2024efficient, casper2024defending}. Moreover, even with the same model, dataset, and attack, evaluations remain difficult to compare due to possible variations in tokenization, chat templates, attention implementations, quantization, fine-tuning methods, etc., that can depend on the chosen library and its version. The combination of high computational demands and vast hyperparameter spaces significantly complicate reliable hyperparameter tuning. Consequently, even when individual experiments are reproducible with provided code, meaningful comparisons between different approaches (e.g., from consecutive papers) become challenging. This makes it more difficult to attribute claimed improvements to methodological changes over subtle variations in experimental setups than it already was in other machine learning domains~\cite{musgrave2020metric, qi2024evaluating}.

\vspace{-5pt}
\subsection{Realigned Objectives}\label{sec:pos-capability-realigned}

We identify a tension between real-world relevance of attack and defense evaluations, manifested by research on proprietary models, and the fundamental need for reproducible and comparable research. Specifically, we note: \textbf{A)} a shift towards black-box evaluations on proprietary models, which makes it more difficult to identify robustness through obscurity and hurts reproducibility (e.g., frequent model updates, limited control). \textbf{B)} Increased evaluation cost, which may result in a lack of third-party evaluations, which would highlight flawed evaluations and approaches and thus establish best practices in the community. \textbf{C)} A lack of community-driven leaderboards makes it difficult to compare approaches across different works. \textbf{D)} Limited comparability between consecutive approaches due to high computational effort and complex LLM pipelines, making it difficult to reliably measure incremental progress.

\textbf{Academia's role in adversarial alignment.} We argue that academia, constrained by limited computational resources, is unlikely to drive state-of-the-art advancements in large-scale alignment techniques. 
Academia's strength lies in exploring foundational ideas in controlled settings to generate reliable results in ``grassroots'' research~\cite{krizhevsky2012imagenet, vaswani2017attention}, whereas industry focuses on scaling the most promising approaches~\cite{brown2020language}. 

Moreover, academia has a unique position in the incentive landscape of AI safety. Similar to independent ``white-hat-hackers'' in cybersecurity~\cite{caldwell2011ethical}, academic researchers have both the freedom and incentive to rigorously test and challenge AI systems. While model providers naturally balance robustness against utility, and safety-focused startups may have commercial interests tied to their defensive approaches, academia faces no such constraints. This synergistic relationship between academia and industry is particularly effective when academic findings are reliable.

Both for the purpose of exploring foundational ideas and leveraging academia's independent incentive structure we argue that academia can maximize its impact by prioritizing the generation of reliable insights over achieving the highest attack performance on the latest proprietary models. This shift requires a focus on open-source, accessible, deterministic, and methodologically transparent defenses, facilitating reproducible findings. 

\textbf{Transparent evaluations.} Addressing \textbf{A)}, we propose two complementary approaches: First, we argue that the rapid advancements in open-source models are gradually reducing the necessity of evaluating proprietary ones. As open-source models continue to achieve state-of-the-art performance, they offer a viable and accessible alternative for robustness research on models with representative capabilities.
In this context, we highlight that robustifying even relatively small open-source models (e.g., those with 7 billion parameters) remains an open research challenge, and we therefore encourage research focused on such models as well. Second, we acknowledge that the robustness of proprietary models has considerable real-world relevance (e.g., assessing the robustness of whole pipelines instead of single models). To foster synergies between academia and industry while ensuring reproducibility, research-focused APIs for LLMs could play a crucial role in ensuring the reliability of evaluations. These APIs could, for example, provide access to fixed-version models for guaranteed periods, along with incentives for academic researchers. Moreover, where feasible, we believe that academia can better support safety research by gaining greater access to information about safety pipelines and training datasets, fostering a more collaborative and transparent approach.


\textbf{Comunity leaderboards and standardized evaluations.} Addressing \textbf{B)}, \textbf{C)} and \textbf{D)}, we propose that researchers and reviewers prioritize comparability and reproducibility over state-of-the-art performance when selecting attacks and defenses. Instead of using the newest proprietary techniques, studies should favor well-documented methods integrated into standardized frameworks. This ensures that evaluations can be reliably reproduced across different works. Reproducibility in LLM research is difficult even for open-source models and code, e.g., due to the internal randomness of flash attention~\cite{maini2024tofu}, making efforts in this direction even more important. To support this, we advocate for the creation of community-driven leaderboards and standardized evaluation frameworks that define clear benchmarks for attacks, defenses, and evaluation pipelines. Reviewers should actively discourage the use of attacks that lack open-source implementations or are not part of established benchmarking frameworks. Similarly, new defenses should not be evaluated solely against the most recent attack but rather against a broad, well-documented set of methods to ensure meaningful comparisons. While some initiatives in this direction exist, they currently cover only a subset of relevant factors (e.g., datasets and attacks). Expanding these efforts to include a wider range of robustness criteria would enhance comparability and encourage research that advances fundamental understanding rather than chasing incremental state-of-the-art improvements.
\begin{tcolorbox}[
    colback=white, colframe=purple!80!black, coltitle=white, fonttitle=\bfseries, 
    rounded corners, enhanced, 
    title=Key Takeaways, 
    attach boxed title to top left={yshift=-2mm, xshift=5mm}, 
    boxed title style={colback=purple!80!black, rounded corners},
    boxsep=0.5mm,
    left=2mm, 
    right=2mm
]
Focus on accessible open-source models and methodologically transparent defenses. Less incentive for proprietary evaluations and breaking the newest models. A community effort towards public leaderboards and standardized evaluation frameworks that prevent robustness through obscurity and enable reproducible and comparable progress.
\end{tcolorbox}




\section{Beyond Discussed Threats}
Most arguments presented in this work regarding emerging challenges in LLM robustness research are applicable beyond the specific threat models discussed as part of the taxonomy. 
These include issues related to complex input-output domains (e.g., precisely defining attack constraints or evaluation goals) and challenges associated with large model sizes, vast number of hyperparameters, and varied implementation settings.
This includes integrity threats, such as model poisoning, as well as attacks on confidentiality (e.g., membership inference, model inversion) and hybrid threats, like model unlearning. 
However, we note that these related areas are likely to present their own unique complexities and challenges within the LLM context that are currently hindering research progress, which requires discussion in future work.
\vspace{-10pt}
\section{Arguments against the Position}

\textbf{Distangeling problems increases the reality gap.} An important counter-argument to our position for disentangling complex problems into well-defined sub-problems is the concern that such an approach may lead to a widening gap between academic research and real-world threats. 

First, real-world adversarial attacks on LLMs (or multimodal models) will rarely target individual models in isolation, but will need to circumvent guardrailes of complex interacting systems with multiple components. This includes input content filters, harmfulness detectors in the output of the model, monitoring of user behavior (e.g., number of send requests), and other pipeline steps. Focusing on simplified proxies may cause researchers to overlook crucial aspects of the problem that only emerge when considering the problem in its full complexity. For instance, by overly focusing on specific types of evasion attacks with constrained input spaces, researchers risk ignoring more sophisticated attacks that exploit system vulnerabilities beyond the model itself~\cite{debenedetti2024privacy}, or target system interfaces in unexpected ways~\cite{carlini2024stealing}. The concern is that simplified settings can create a false sense of security, where progress is made on the auxiliary task but systems are still vulnerable to novel real-world attacks. 

For example, in image classifications, defenses that show strong performance against $\ell_p$ norm attacks in controlled settings may fail against unforeseen threats, such as simple color changes~\cite{laidlaw2019functional}, whether conditions~\cite{kaufmann2019testing}, adversarial patches or objects~\cite{song2018physical, athalye2018synthesizing}. Moreover, in the context of LLMs, recent work demonstrated that simply reformulating malicious requests in past tense~\cite{andriushchenko2024does} or in a polite manner~\cite{xhonneux2024efficient} is sufficient to jailbreak most state-of-the-art models. These weaknesses were not revealed by automatic adversarial attacks and similar but through human effort and similar findings require researchers to study the real-world problem directly, without simplifications. 

Another argument against a focus on well-defined sub-problems is the broad attack surface of foundation models from chat assistants to autonomous agents. The majority of threats related to unexplored and upcoming applications will not be formally well-defined. Here, research requires identifying novel threat surfaces in new application areas. This kind of research has been proven to be highly relevant in the past, including the identification of vulnerabilities in commercial hashing algorithms~\cite{levenson2021apple}, seemingly reliable protections of art against generative AI~\cite{honig2024adversarial}, or vulnerabilities of newly developed benchmarks, such as LLM leaderboard~\cite{huang2025exploring}.

\textbf{New opportunities for defenses.} In this work, we primarily emphasize the increased complexity of achieving robustness in LLMs compared to previous research domains. However, the semantic output and input space of LLMs not only complicate measuring robustness but also open up new opportunities for defenses. Early adversarial detection methods in computer vision focused on identifying \textit{small}, \textit{imperceptible} norm-bounded perturbations, proved largely ineffective~\cite{carlini2017adversarial}. Moreover, detection in the output space was not possible due to the prevalence of simple classification tasks. Conversely, recent advancements in the LLM domain demonstrate the strong empirical results of input and output classifiers for defending against harmful requests~\cite{kim2023robust, sharma2025constitutional}. This suggests that the inherent semantics of the LLMs domain, while presenting new challenges, also facilitate the development of new defense mechanisms previously unavailable in other domains.

\textbf{Capability progress is too fast.} Another argument against the proposed position of reliable research in constrained settings is the rapid pace of progress in model capabilities. By simplifying the problem, researchers risk creating solutions that will quickly become obsolete as AI models evolve. This could be specifically the case as current adversarial attack algorithms often rely on LLMs to jailbreak other models. Rather than trying to solve isolated, small problems in the field, researchers should concentrate on large, fundamental questions that are not dependent on current technical limitations, as they may be resolved with progress in LLM capability. Moreover, while open-source models are increasing in their capabilities, research on open-source models may always lag behind proprietary ones and academia may recover mistakes already identified by industry labs. 

\vspace{-10pt}
\section{Conclusion}


In this position paper, we argue that current objectives in adversarial alignment for LLMs are misaligned, hindering research progress.\footnote{The goal of this position paper is to identify misaligned objectives in \textit{academia}. Our intent is not to offer a political stance on the risks associated with LLMs nor to critique the role of industry.} 
We provide concrete examples of historical challenges that have slowed adversarial robustness research and identify novel issues specific to LLMs. To address these challenges, we propose reducing problem complexity by disentangling adversarial alignment into distinct sub-problems focusing on LLM robustness while reducing ambiguities related to the evaluation of alignment.
Moreover, we emphasize how academia can contribute more effectively to this industry-driven field by prioritizing rigorous research that advances fundamental understanding and generates new insights rather than merely optimizing for performance on the latest model benchmarks. 

%

\section*{Impact Statement}
Adversarial attacks on LLMs can have considerable consequences in real-world applications. Still, as machine-learning robustness has been an unsolved research problem for the last decade, we believe that the best way to approach this problem is through culminating awareness. Currently, it seems unlikely that the robustness issue can be completely resolved through technical means. Thus, making people aware of the harmful use cases and limitations of these models appears to be necessary to avoid irresponsible deployment of such models for critical applications and to reduce the harm malicious actors can cause. Moreover, this work does not introduce new technical innovations. Instead, it addresses impediments in robustness research and explores potential solutions to accelerate research progress.
\clearpage

\bibliography{references}
\bibliographystyle{icml2025}

\newpage
\appendix
\onecolumn

\setlength{\myboxwidth}{13cm}

\end{document}